\documentclass[letterpaper]{article} 
\usepackage{aaai2026}  
\usepackage{times}  
\usepackage{helvet}  
\usepackage{courier}  
\usepackage[hyphens]{url}  
\usepackage{graphicx} 
\usepackage{natbib}  
\usepackage{caption} 
\usepackage{algorithm}
\usepackage{amsmath} 
\usepackage{multirow} 
\usepackage[table,xcdraw]{xcolor}
\usepackage{amssymb} 
\usepackage{xcolor}
\usepackage{tabularx} 
\usepackage{graphicx} 
\usepackage{array}
\usepackage{algorithmic}

\usepackage{newfloat}
\usepackage{listings}
\DeclareCaptionStyle{ruled}{labelfont=normalfont,labelsep=colon,strut=off} 
\floatstyle{ruled}
\newfloat{listing}{tb}{lst}{}
\floatname{listing}{Listing}
\title{UniABG: Unified Adversarial View Bridging and Graph Correspondence for Unsupervised Cross-View Geo-Localization}
\author {
    Cuiqun Chen\textsuperscript{\rm 1},
    Qi Chen\textsuperscript{\rm 1},
    Bin Yang\textsuperscript{\rm 2}\thanks{Corresponding Author},
    Xingyi Zhang\textsuperscript{\rm 1*}
}
\affiliations {
    \textsuperscript{\rm 1}School of Computer Science and Technology, Anhui University, China\\
    \textsuperscript{\rm 2}  School of Computer Science, Wuhan University, Wuhan, China\\
    chencuiqun@ahu.edu.cn, cq@stu.ahu.edu.cn, yangbin\_cv@whu.edu.cn, xyzhanghust@gmail.com
}

\usepackage{bibentry}

\begin{document}

\maketitle

\begin{abstract}
Cross-view geo-localization (CVGL) matches query images ($\textit{e.g.}$, drone) to geographically corresponding opposite-view imagery ($\textit{e.g.}$, satellite). While supervised methods achieve strong performance, their reliance on extensive pairwise annotations limits scalability. Unsupervised alternatives avoid annotation costs but suffer from noisy pseudo-labels due to intrinsic cross-view domain gaps. To address these limitations, we propose $\textit{UniABG}$, a novel dual-stage unsupervised cross-view geo-localization framework integrating adversarial view bridging with graph-based correspondence calibration. Our approach first employs View-Aware Adversarial Bridging (VAAB) to model view-invariant features and enhance pseudo-label robustness. Subsequently, Heterogeneous Graph Filtering Calibration (HGFC) refines cross-view associations by constructing dual inter-view structure graphs, achieving reliable view correspondence. Extensive experiments demonstrate state-of-the-art unsupervised performance, showing that UniABG improves Satellite $\rightarrow$ Drone AP by +10.63\% on University-1652 and +16.73\% on SUES-200, even surpassing supervised baselines. The source code is available at https://github.com/chenqi142/UniABG
\end{abstract}

\section{Introduction}
Cross-view geo-localization (CVGL) addresses the critical task of determining geographic coordinates for ground-level or aerial query images by matching them against georeferenced satellite imagery. The proliferation of unmanned aerial vehicles (UAVs) has significantly expanded CVGL's scope, enabling centimeter-level positioning essential for urban navigation, autonomous systems, and augmented reality. UAVs' flexible low-altitude imaging capability renders them particularly advantageous for fine-grained localization in complex environments. 

Nevertheless, CVGL faces inherent challenges stemming from extreme cross-modal viewpoint and appearance variations. Prior efforts mitigate these through geometry-aware architectures~\cite{regmi2019bridging, liu2019lending}, semantic-guided embedding~\cite{hu2018cvm, zhao2024transfg}, and contrastive learning frameworks~\cite{cai2019ground, deuser2023sample4geo}. While achieving notable performance, such supervised methods rely extensively on large-scale manually annotated cross-view image pairs, which incur prohibitive labelling costs.

\begin{figure}[t] 
    \centering 
    \includegraphics[width=0.45\textwidth]{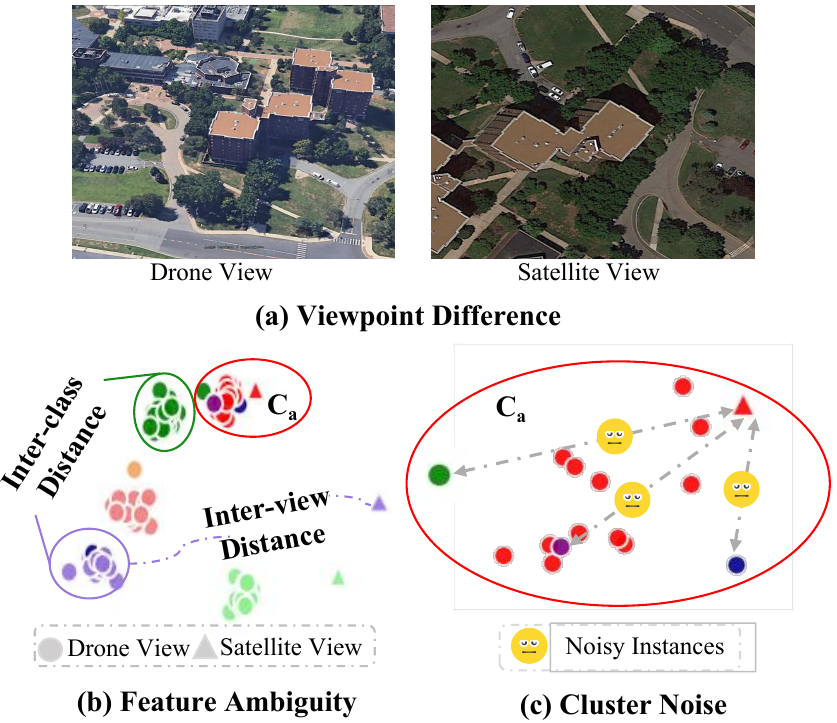} 
    \caption{Key challenges in unsupervised CVGL. (a) Drastic appearance differences between drone and satellite views. (b) Feature space ambiguity, where the distance between different views of the same category may exceed the distance within different categories. (c) An enlarged view of the clustering space of category A ($\mathbf{C_a}$). Noisy instances within clusters leading to incorrect pseudo-label association. \textbf{Different colors represent different categories.}} 
    \label{fig:motivation} 
\end{figure}

To circumvent this limitation, recent research has shifted toward unsupervised cross-view geo-localization (UCVGL). Pioneering works include Li et al.~\cite{li2024unleashing}, who synthesize projected ground images aligned with satellite views to generate pseudo-pairs, and self-supervised approaches~\cite{li2024learning, huempowering, wang2025tokenmatcher} that leverage foundation models with Expectation-Maximization (EM) mechanisms and consistency regularization for pseudo-label assignment. Wang et al.~\cite{wang2025coarse, adca, yang2023dual} further adopt clustering-based contrastive learning to extract instance- and centroid-level features without supervision.

Despite these advances, the current UCVGL paradigm performs cross-view label association directly during training, overlooking two critical limitations.
As shown in Fig.~\ref{fig:motivation}(a-b), divergent imaging altitudes and resolutions between drone/satellite modalities cause identical categories to exhibit larger feature-space distances across views than different categories within the same view. This view distribution misalignment propagates clustering errors through erroneous label associations. Furthermore, existing UCVGL methods employ simplistic label transfer mechanisms that fail under cluster impurity. Fig.~\ref{fig:motivation}(c) demonstrates how noisy instances trigger catastrophic association errors when directly mapped across views. 

To address these challenges, we propose \textbf{UniABG}, a dual-stage framework for unsupervised cross-view geo-localization. Our approach synergizes adversarial view bridging with graph-based correspondence calibration, mitigating feature-level view gaps while establishing noise-robust cross-view correspondences. Through this dual-stage design, UniABG concurrently generates high-fidelity association pairs and learns view-consistent representations, achieving significant performance gains on standard unsupervised CVGL benchmarks.

In the first stage, to bridge the inherent geometric and spectral divergence between drone and satellite views, we establish a unified embedding space via View-Aware Adversarial Bridging (VAAB). This framework deploys feature extractors that compete adversarially against a domain classifier. By deliberately confusing the classifier across all three views – drone, satellite, and an Auxiliary Pseudo View (APV) – this optimization eliminates view-specific artifacts and forces discriminative, view-invariant feature learning. The APV is generated through cross-view style transfer between existing drone/satellite data. This synthetic view preserves structural semantics while simulating perspective transitions from low-altitude drone imagery to nadir-aligned satellite views, serving as a geometric intermediary that enables progressive viewpoint adaptation.

The second stage addresses error propagation caused by distorted neighborhood graph filtering structures during early clustering. Our heterogeneous graph filtering calibration (HGFC) module constructs dual graphs encoding inter-view structures to achieve robust correspondence matching. Specifically, for correspondence purification, we introduce mutual k-reciprocal neighbour filtering, which requires satellite candidates to neighbour both the drone image and its APV in the heterogeneous graphs. This symmetric constraint preserves geometrically consistent pairs while eliminating ambiguous matches from noisy clusters, thereby resolving the problematic noisy instances.

In summary, our principal contributions encompass:

(1) We propose the first dual-stage framework, \emph{e.g.,} UniABG, for unsupervised cross-view geo-localization that integrates adversarial learning with graph correspondence filtering, jointly addressing the critical challenges of view discrepancy and association noise.

(2) We introduce a view-aware adversarial bridging strategy to mitigate cross-view discrepancies. This approach combines an auxiliary pseudo-view to model view-invariant embedding spaces while simultaneously enhancing cross-view feature discriminability.

(3) We develop a robust correspondence calibration mechanism through heterogeneous graph construction and mutual k-reciprocal neighbour filtering, significantly enhancing matching accuracy by eliminating ambiguous cross-view associations while preserving geometrically consistent pairs.

(4) We demonstrate state-of-the-art performance on University-1652 and SUES-200 benchmarks, where our approach outperforms all existing unsupervised methods and surpasses most supervised baselines, establishing new standards for unsupervised cross-view geo-localization.

\section{Related Work}
\subsection{Supervised Cross-View Geo-localization}
Cross-View Geo-Localization (CVGL) is critical for tasks such as autonomous navigation and augmented reality, where large viewpoint differences between ground and aerial imagery pose significant challenges. Early works focused on global feature learning, such as CVM-Net \cite{tian2017cross} and spatial-aware aggregation \cite{shi2019spatial}, while orientation priors \cite{liu2019lending} helped reduce spatial ambiguity. Subsequent studies addressed geometric misalignment, including Optimal Feature Transport (OFT) \cite{shi2020optimal} and disentangled geometric learning \cite{zhang2023cross_aaai}.
Recent methods emphasize semantic reasoning and attention-based models to improve robustness. Geo-Net \cite{zhu2021geographic} and SemGeo \cite{rodrigues2023semgeo} leverage semantic cues, while Transformer-based approaches such as TransGeo \cite{zhu2022transgeo} and layer-wise correlation \cite{yang2021cross} enhance long-range dependencies. Other strategies incorporate part-level correlation \cite{wang2021each} and multi-candidate matching \cite{zhu2021vigor} to improve performance.

\subsection{Unsupervised Cross-View Geo-localization}
The heavy reliance of supervised CVGL methods on labelled image pairs has motivated the development of unsupervised approaches. Li \emph{et al}. \cite{li2024unleashing} proposed a framework that generates pseudo-labels through cross-view projection, while Li \emph{et al}. \cite{li2024learning} introduced an EM-based self-supervised scheme but suffered from noisy label propagation. To alleviate such issues, Wang \emph{et al}. \cite{wang2025coarse} employed clustering contrastive learning to learn cross-view representations without annotations, though it remained vulnerable to error accumulation. Building on these insights, we design a dual-stage framework that stabilizes clustering before performing cross-view association and integrates graph-based filtering to ensure contrastive supervision.


\begin{figure*}[t] 
    \centering 
    \includegraphics[width=0.85\textwidth]{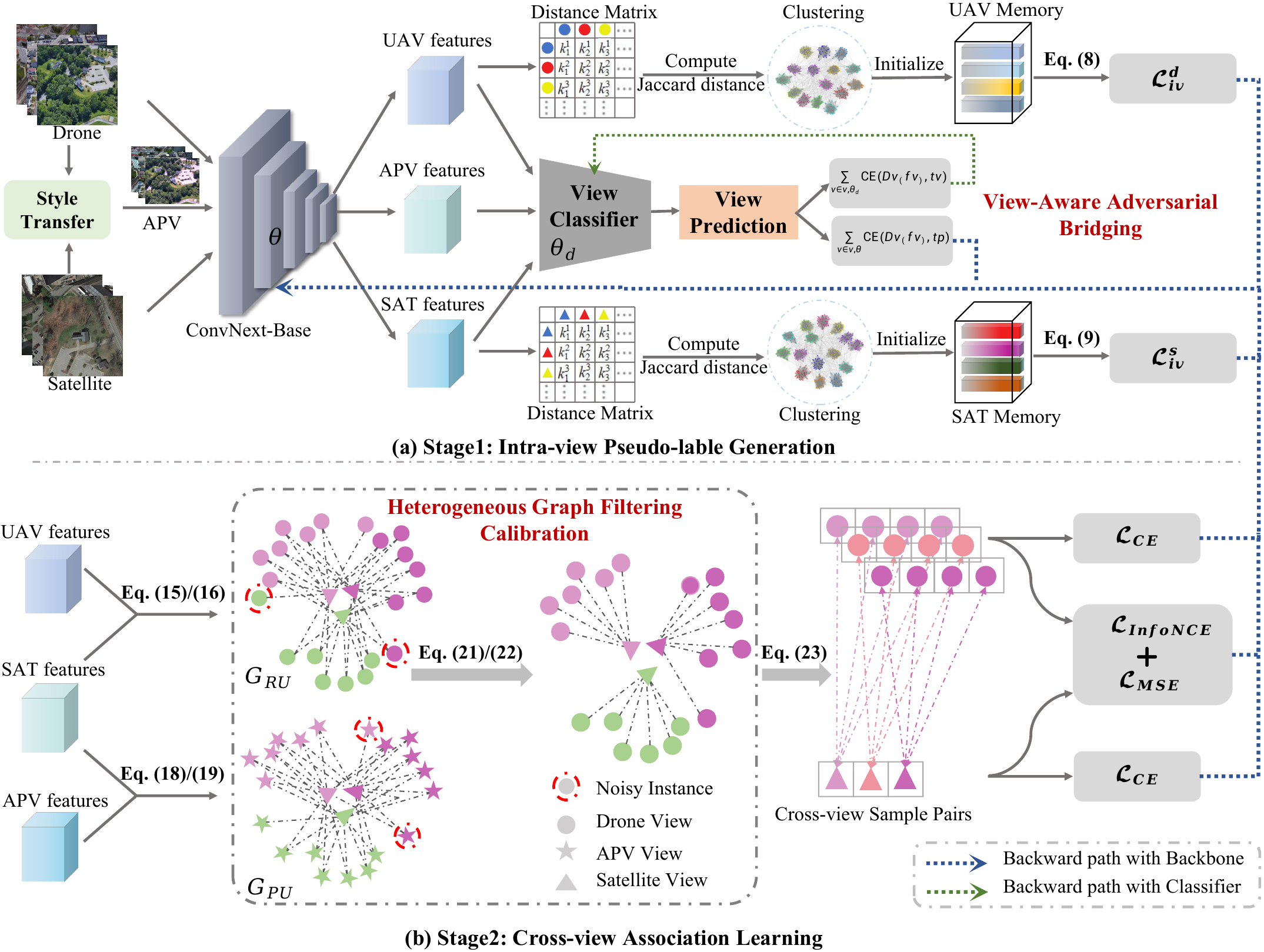} 
    \caption{The overall architecture of our proposed UniABG is a dual-stage model. The first stage employs adversarial learning to reduce the differences between perspectives. The second stage constructs cross-view association data through heterogeneous graph filtering calibration for supervised learning (for more details, please refer to the text).} 
    \label{fig:UniABG} 
\end{figure*}

\section{Method}
This section presents UniABG, our proposed dual-stage framework for cross-view matching (Fig. \ref{fig:UniABG}). In the first stage, pseudo-labels are generated via clustering and optimized through intra-view contrastive learning. These pseudo-labels are further refined using view-aware adversarial learning to bridge the cross-view discrepancy. Subsequently, heterogeneous graph filtering calibration refines cross-view correspondences, effectively enhancing matching reliability. UniABG achieves state-of-the-art performance on the University-1652 and SUES-200 datasets.

\textbf{Problem Formulation.} In the UCVGL, given an unlabeled drone-satellite image set, denoted as \(\{X_d, X_s\}\), where $X^d = \left\{ x_i^d \middle| i = 1, 2, \ldots, N^d \right\}$ represents the input drone image and $X^s = \left\{ x_i^s \middle| i = 1, 2, \ldots, N^s \right\}$ represents the input satellite image. $N^d$ and $N^s$ represent the number of images in two views, respectively. The goal is to mine reliable positive and negative samples under this condition to complete the UCVGL task.

\subsection{Dual-stage Unsupervised Cross-View Geo-Localization Baseline}
Recent advances in cluster-based contrastive learning demonstrate significant potential for uncovering inherent data structures without labels. Building on this, we propose a dual-stage baseline for unsupervised cross-view geo-localization. Our framework employs ConvNeXt-B \cite{xia2024enhancing} for feature extraction and operates as follows: \emph{(1) Stage 1: Intra-view Pseudo-label Generation.} Clustering methods generate pseudo-labels within each view. To optimize feature discrimination, we apply intra-view contrastive learning constrained by a dual-contrastive strategy \cite{yang2022augmented}. \emph{(2) Stage 2: Cross-view Association Learning.} Cross-view data associations are constructed based on cosine similarity. These associations then drive cross-view contrastive learning to align view representations.

\textbf{Intra-view Pseudo-label Generation.} We first construct representative feature banks (memory dictionaries) by clustering unlabeled data within each view, serving as anchors for subsequent contrastive learning. Specifically, we extract features from drone ($X^d$) and satellite ($X^s$) images using a shared backbone network $\mathcal{F}_{\text{backbone}}$, yielding cross-view features
\begin{equation}
    f_i^d = \mathcal{F}_{\text{backbone}}(x_i^d), i \in \{1, 2, \ldots, N^d\},
\end{equation}
\begin{equation}
    f_i^s = \mathcal{F}_{\text{backbone}}(x_i^s), i \in \{1, 2, \ldots, N^s\},
\end{equation}
where $F_d = \left\{ f_i^d \right\}_{i=1}^{N^d}$ and $F_s = \left\{ f_i^s \right\}_{i=1}^{N^s}$ represent the feature sets of two views, respectively. To discover inherent data structures without labels, we apply DBSCAN \cite{ester1996density} separately on $F_d$ and $F_s$ to generate view-specific pseudo-labels
\begin{equation}
\hat{Y}^d = \text{DBSCAN}(F_d),
\end{equation}
\begin{equation}
\hat{Y}^s = \text{DBSCAN}(F_s),
\end{equation}
where $\hat{Y}^d = \{\hat{y}_1^d, \hat{y}_2^d, \ldots, \hat{y}_{N^d}^d\}$ and $\hat{Y}^s = \{\hat{y}_1^s, \hat{y}_2^s, \ldots, \hat{y}_{N^s}^s\}$ denote the pseudo labels of the input images $X^d$ and $X^s$. Using these pseudo-labels, we compute the cluster centroids (prototypes) $\phi_k^d$ for drone view and $\phi_l^s$ for satellite view as the mean feature within each cluster
\begin{equation}
\phi_k^d = \frac{1}{|\mathcal{H}_k^d|} \sum_{\mathbf{f}_n^d \in \mathcal{H}_k^d} \mathbf{f}_n^d, \quad
\phi_l^s = \frac{1}{|\mathcal{H}_l^s|} \sum_{\mathbf{f}_n^s \in \mathcal{H}_l^s} \mathbf{f}_n^s,
\end{equation}
where $\mathcal{H}_k^d = { \mathbf{f}_n^d \mid \hat{y}_n^d = k }$ is the set of drone features belonging to cluster $k$, $\mathcal{H}_l^s = { \mathbf{f}_m^s \mid \hat{y}_m^s = l }$ is the set of satellite features belonging to cluster $l$, and $|\cdot|$ denotes set cardinality.
Finally, we initialize the drone memory dictionary $\mathcal{M}_d$ and satellite memory dictionary $\mathcal{M}_s$ with their respective cluster prototypes
\begin{equation}
\mathcal{M}_d \leftarrow \{\phi_1^d, \phi_2^d, \cdots, \phi_K^d\}, \quad
\mathcal{M}_s \leftarrow \{\phi_1^s, \phi_2^s, \cdots, \phi_L^s\}.
\end{equation}

To further enhance the quality of intra-view pseudo-label generation, we employ an intra-view contrastive loss as a regularization mechanism. This approach optimizes the memory space distribution structure by reducing intra-memory feature distances while increasing inter-memory separation. 
Given drone and satellite query features $q^d$ and $q^s$, we compute the contrastive loss for drone view and satellite view by the following equations
\begin{equation}
\mathcal{L}_{iv} = \mathcal{L}_{iv}^d + \mathcal{L}_{iv}^s,
\end{equation}
\begin{equation}
\mathcal{L}_{iv}^d = -\log \frac{\exp(q^d \cdot \phi_+^d / \tau)}{\sum_{k=0}^K \exp(q^d \cdot \phi_k^d / \tau)},
\end{equation}
\begin{equation}
\mathcal{L}_{iv}^s = -\log \frac{\exp(q^s \cdot \phi_+^s / \tau)}{\sum_{l=0}^L \exp(q^s \cdot \phi_l^s / \tau)},
\end{equation}
where $\phi_+$ is the positive memory corresponding to the pseudo label of $q$ and the $\tau$ is a temperature. $\phi_{k}^{d}$ and $\phi_{l}^{s}$ represent the view cluster center features, respectively.

\textbf{Cross-view Association Learning.} 
Following the initial feature extraction, we perform the cross-view association learning for view matching. The objective here is to establish direct correspondences between individual drone and satellite images using generated pseudo-labels for supervised contrastive learning.

Given the feature sets for the drone view, \(F_d = \{f_i^d\}_{i=1}^{N^d}\), and the satellite view, \(F_s = \{f_j^s\}_{j=1}^{N^s}\), we construct cross-view associations using a greedy nearest-neighbour strategy. For each drone instance feature \(f_i^d\), we compute its cosine similarity with every satellite instance feature in the set \(F_s\). The satellite instance that yields the highest similarity is designated as the positive pair for \(f_i^d\). This process identifies a corresponding satellite instance index \(j^*\) for each drone instance \(i\)
\begin{equation}
j^* = \underset{j \in \{1, \dots, N^s\}}{\operatorname{argmax}} \left( \frac{(f_i^d)^\top f_j^s}{\|f_i^d\| \|f_j^s\|} \right).
\label{eq:instance_association}
\end{equation}

Once the positive drone-satellite pair \((f_i^d, f_{j^*}^s)\) is identified via greedy matching (Eq. \ref{eq:instance_association}), all other satellite instances \(\{f_j^s \mid j \neq j^*\}\) are treated as negatives for the drone query \(f_i^d\). This allows us to define a cross-view contrastive learning objective: pull the positive pair closer in the embedding space while pushing negatives apart. We optimize the following supervised loss \cite{xia2024enhancing}:
\begin{equation}
\mathcal{L}_{\text{sup}} = \mathcal{L}_{\text{InfoNCE}} + \mathcal{L}_{\text{MSE}} + \mathcal{L}_{\text{CE}},
\label{eq:dac_loss}
\end{equation}
where $\mathcal{L}_{\text{InfoNCE}}$ learns scene-discriminative features, $\mathcal{L}_{\text{MSE}}$ establishes cross-view correspondences, and $\mathcal{L}_{\text{CE}}$ optimizes positive sample representations. However, this approach is critically vulnerable to noise in the initial association. Greedy matching is highly susceptible to visual ambiguities, generating erroneous positive pairs. Propagating these incorrect pseudo-labels into the supervised learning framework leads to significant error accumulation, thereby fundamentally limiting model performance. This underscores the necessity of our bridging and correspondence purification strategy in \emph{UniABG}, which refines pseudo-labels and enhances cross-view robustness.

\subsection{View-Aware Adversarial Bridging}
\label{subsec:adversarial_bridging}
A fundamental challenge in unsupervised cross-view geo-localization (UCVGL) stems from the significant domain gap induced by drastic perspective differences between drone and satellite imagery. This discrepancy, characterised by extreme scale variations, geometric distortions, and illumination disparities, results in substantial misalignment of feature distributions across views\cite{liu2024learning}. \textit{This misalignment is particularly detrimental in the unsupervised setting}. However, naive cross-view clustering assumes cross-view feature compatibility but fails to bridge domain gaps, yielding performance-limiting noisy pseudo-labels.

To explicitly mitigate this cross-view distribution divergence for robust clustering, we propose a view-aware adversarial bridging (VAAB) module. 
Inspired by cross-modal shared representation learning \cite{linMITMLcvpr, yang2025progressive, Yang_2023_ICCV, yang2023translation, yang2024shallow, yao2025unsupervised}, VAAB employs a synthetic pseudo-view as a transitional domain with adversarial constraints. 
Its core mechanism uses triplet-view adversarial training to reduce view-specific characteristics, learning discriminative cross-view representations.

\textbf{Auxiliary Pseudo View.} Referring to \cite{reinhard2002color, wang2025coarse}, we obtain the auxiliary pseudo view (APV) by performing a global color transfer from the satellite domain to the drone domain. Both images are first converted into the Lab color space, and then channel-wise statistics (mean $\mu_c$ and standard deviation $\sigma_c$ for $c \in \{L, a, b\}$) are computed.  For the satellite domain, we compute the global mean and standard deviation ($\mu^s_c$, $\sigma^s_c$) for each channel $c \in \{L, a, b\}$) over the entire dataset. For each input drone image, we compute its individual statistics ($\mu^d_c$, $\sigma^d_c$) and apply the following channel-wise transformation
\begin{equation}
l'_c = \frac{\sigma^s_c}{\sigma^d_c}(l_c - \mu^d_c) + \mu^s_c,
\end{equation}
where $l_c$ and $l'_c$ denote the original and transformed pixel values, and the superscripts $s$ and $d$ indicate satellite and drone domains, respectively. This transformation standardizes drone image appearance while preserving its structural content, forming the auxiliary pseudo view.

\textbf{Adversarial View Bridging.} While style transfer provides pixel-level approximation, it fails to ensure semantic-level feature consistency across views. To achieve deeper domain invariance essential for cross-view association, we introduce an adversarial view bridging strategy. This approach transcends conventional feature alignment by explicitly harmonizing feature distributions through adversarial training. The core mechanism employs a view discriminator $D_v$ that attempts to classify feature origins (drone $x^d$, satellite $x^s$, or APV $x^p$). By adversarially training the backbone to confound $D_v$, we force it to suppress view-specific artefacts and extract viewpoint-invariant representations.

Formally, features from all views are extracted via shared backbone $F_{\text{B}}$:
\begin{align}
f^d_i = F_{\text{B}}(x^d_i), \quad
f^s_i = F_{\text{B}}(x^s_i), \quad
f^p_i = F_{\text{B}}(x^p_i).
\end{align}
A view discriminator $D_v$ is trained to identify perspective origins, while $F_{\text{B}}$ adversarially learns to induce geometric confusion – making features indistinguishable across views yet discriminative for location. This is achieved through a unified adversarial objective
\begin{equation}
\mathcal{L}_{\text{VAAB}} = {\sum_{v \in \mathcal{V}, \theta_d} \text{CE}\big(D_v(f^v), t^v\big)} + {\sum_{v \in \mathcal{V}, \theta} \text{CE}\big(D_v(f^v), t^p\big)},
\end{equation}
where $\mathcal{V} = {d, s, p}$ denotes drone/satellite/APV views. Thus, $t^d$, $t^s$, and $t^p$ are the corresponding view labels for three views. $\theta_d$ denotes the parameters of the view classifier and $\theta$ denotes the parameters of the backbone. This convergence emerges from complementary roles where the first item of $\mathcal{L}_{\text{VAAB}}$ forces the discriminator to explore view-specific patterns, while the second drives the backbone to eliminate spectral and geometric biases. This establishes a foundation for reliable pseudo-label generation in subsequent clustering and correspondence optimization stages. 

\subsection{Heterogeneous Graph Filtering Calibration} 

The quality of cross-view association data is crucial to the effectiveness of subsequent supervised training. However, traditional methods mostly rely solely on the feature similarity within a single view to perform matching. This approach is highly sensitive to noise and is prone to generating a large number of ambiguous matches, making it difficult to ensure the quality of the association data\cite{xu2025efficient}.

To address this, we propose a Heterogeneous Graph Filtering Calibration (HGFC) module based on the auxiliary pseudo-view. HGFC aims to construct high-quality cross-view association data by exploiting the structural consensus across two complementary views: drone-to-satellite and APV-to-satellite. Our key hypothesis is that true matching relationships should exhibit structural consistency across multiple feature manifolds, thereby effectively filtering out incorrect matches and improving the accuracy and reliability of the associations.

\textbf{Heterogeneous Graph Construction.} Given the extracted feature sets of drone and satellite images, denoted as $F_d = \{f^d_i\}$ and $F_s = \{f^s_j\}$ respectively, we compute the cosine similarity between each drone feature and all satellite features
\begin{equation}
\text{sim}(f^d_i, f^s_j) = \frac{(f^d_i)^\top f^s_j}{\|f^d_i\| \cdot \|f^s_j\|}.
\end{equation}

Then, for each $f^d_i$, we construct its $k$-nearest satellite neighbors
\begin{equation}
N_k^{RU}(f^d_i) = \underset{f^s_j \in F_s}{\text{Top-}k} \ \text{sim}(f^d_i, f^s_j),
\end{equation}
which defines the edge set $\mathcal{E}_{RU}$ of the \emph{Real-to-Real Graph} $G_{RU} = (F_d, F_s, \mathcal{E}_{RU})$.
The edge set $\mathcal{E}$ is defined as the connection relationships based on feature similarity
\begin{equation}
\mathcal{E}_{RU} = \{(f^d, f^s) \mid f^s \in N_k(f^d)\}.
\end{equation}

Similarly, we use the auxiliary pseudo view feature set $F_p = \{f^p_i\}$ and compute its similarity with $F_s$
\begin{equation}
\text{sim}(f^p_i, f^s_j) = \frac{(f^p_i)^\top f^s_j}{\|f^p_i\| \cdot \|f^s_j\|},
\end{equation}
and define its neighbour set as
\begin{equation}
N_k^{PU}(f^p_i) = \underset{f^s_j \in F_s}{\text{Top-}k} \ \text{sim}(f^p_i, f^s_j),
\end{equation}
which forms the \emph{Pseudo-to-Real Graph} $G_{PU} = (F_p, F_s, \mathcal{E}_{PU})$.
The edge set $\mathcal{E}$ is defined as the connection relationships based on feature similarity
\begin{equation}
\mathcal{E}_{PU} = \{(f^p, f^s) \mid f^s \in N_k(f^p)\}.
\end{equation}

\textbf{Topological Consistency Alignment.} To filter out noisy matches, we align the graph structures across the two views. Specifically, we apply mutual $k$-nearest neighbour (mutual-KNN) verification and evaluate the consistency of neighbours between the two graphs. For a satellite feature $f^s_j$, let $N_{RU}(f^s_j)$ and $N_{PU}(f^s_j)$ denote the sets of UAV and pseudo-view features that link to $f^s_j$ in the respective graphs. The cross-graph consistency score is computed as
\begin{equation}
s_{ij}^{\text{cross}} = \frac{|N_k^{RU}(f^s_j) \cap N_k^{PU}(f^s_j)|}{k},
\end{equation}
and only satellite features with $s_{ij}^{\text{cross}} > \tau$ are retained for reliable association.

\textbf{Semantics-Guided Intra-Cluster Weighted Voting.} To suppress outliers and leverage the collective evidence within each cluster, we introduce a semantically guided weighted voting mechanism that uses semantic similarity and structural confidence to refine pseudo-labels. For a candidate pair $(f^d_i, f^s_j)$, the weighted confidence is defined as
\begin{equation}
\omega_{ij} = \text{sim}(f^d_i, f^s_j) \cdot s_{ij}^{\text{cross}}.
\end{equation}
Within each UAV cluster $\mathcal{C}_k$, the final association is obtained via
\begin{equation}
\hat{y}_i = \arg\max_c \sum_{f^s_j \in \mathcal{C}_c} \omega_{ij},
\end{equation}
where $\hat{y}_i$ is the refined pseudo-label, and $\mathcal{C}_c$ denotes the $c$-th cluster of satellite features. This heterogeneous graph filtering calibration ensures that associations are not only based on appearance similarity but also structurally validated across heterogeneous views, thus significantly improving the precision of cross-view matching under unsupervised settings.

\subsection{The Overall Loss Function} 
UniABG employs a dual-stage optimization strategy. The first stage leverages view-aware adversarial learning to reduce cross-view discrepancies and extract discriminative features, facilitating robust clustering. The stage 1 objective combines
\begin{equation}
\mathcal{L}_{\text{stage1}} = \mathcal{L}_{\text{iv}} + \lambda \cdot \mathcal{L}_{\text{VAAB}},
\end{equation}
where $\mathcal{L}_{\text{iv}}$ enforces intra-view feature consistency, and $\mathcal{L}_{\text{VAAB}}$ ensures cross-view domain invariance. These losses jointly learn discriminative, view-invariant representations. 

Using heterogeneous graph filtering, we generate high-confidence matching pairs from the clustered view images. These purified pairs serve as supervisory signals for the second stage, optimized via supervised contrastive loss $\mathcal{L}_{\text{sup}}$.

\section{Experiments}
\subsection{Datasets and Experimental Settings}
We evaluate our method on two public cross-view geo-localization benchmarks: University-1652 \cite{zheng2020university} and SUES-200 \cite{zhu2023sues} using Recall@K (R@K) and Average Precision (AP).

\textbf{University-1652} \cite{zheng2020university} contains 1,652 university buildings captured from satellite, ground-level, and synthetic drone perspectives. This benchmark supports two tasks: 1) \textbf{Satellite $\rightarrow$ Drone} uses synthetic drone images as queries to retrieve matching satellite images, and 2) \textbf{Drone $\rightarrow$ Satellite} uses satellite images as queries for drone navigation in synthetic drone-view galleries. The dataset provides per-location imagery, including 54 synthetic drone images, 1 satellite image, and multiple ground images across 72 universities.

\textbf{SUES-200} \cite{zhu2023sues} introduces drone imagery captured at four heights (150m, 200m, 250m, 300m) paired with satellite views. It addresses altitude variations across diverse scenes (parks, schools, lakes). Supported tasks include: 1) \textbf{Satellite $\rightarrow$ Drone} where drone images (all heights) query satellite galleries, and 2) \textbf{Drone $\rightarrow$ Satellite} where satellite images query drone navigation. This dataset contains 24,120 drone images with corresponding satellite imagery across 200 scenes.

\textbf{Implementation Details}. 
We implement our framework in PyTorch and conduct training on four NVIDIA RTX 4090 GPUs (24GB each). 
All input images are resized to $384\times384$ pixels. 
We adopt ConvNeXt-Base and a supervised objective as baseline \cite{xia2024enhancing}. 
Both training stages use a batch size of 24 (12 drone + 12 satellite image pairs per GPU) for 5 epochs. 
Optimization employs AdamW with an initial learning rate of $1\text{e}^{-3}$, following a cosine decay schedule. 
The hyperparameter $\lambda$ is fixed at 0.1.

\begin{table}[t]
    \centering
    \caption{The results of ablation studies of the proposed method on the University-1652. B represents the baseline.}
    \label{tab:ablation_results}
     \setlength{\tabcolsep}{1mm}
    {
    \begin{tabular}{lll|cccc}
\hline
\multicolumn{3}{l|}{\multirow{2}{*}{Method}}         & \multicolumn{2}{c}{Drone $\rightarrow$ Satellite} & \multicolumn{2}{c}{Satellite $\rightarrow$ Drone} \\
\multicolumn{3}{l|}{}                                & R@1               & AP                & R@1               & AP                \\ \hline
\multicolumn{3}{l|}{B}                              & 35.94             & 41.64             & 65.47             & 35.61             \\ \hline
\multicolumn{3}{l|}{\multirow{2}{*}{B+VAAB}}       & 60.36             & 65.03             & 80.74             & 58.77             \\
\multicolumn{3}{l|}{}                                & $\textcolor{red}{\uparrow} \scriptsize{ \textcolor{red}{24.42}}$             & $\textcolor{red}{\uparrow} \scriptsize{ \textcolor{red}{23.57}}$             & $\textcolor{red}{\uparrow} \scriptsize{ \textcolor{red}{15.27}}$             & $\textcolor{red}{\uparrow} \scriptsize{ \textcolor{red}{23.16}}$             \\ \hline
\multicolumn{3}{l|}{\multirow{2}{*}{B+HGFC}}       & 90.83             & 92.85             & 94.57             & 90.87             \\
\multicolumn{3}{l|}{}                                & $\textcolor{red}{\uparrow} \scriptsize{ \textcolor{red}{54.89}}$             & $\textcolor{red}{\uparrow} \scriptsize{ \textcolor{red}{51.21}}$             & $\textcolor{red}{\uparrow} \scriptsize{ \textcolor{red}{29.10}}$             & $\textcolor{red}{\uparrow} \scriptsize{ \textcolor{red}{55.26}}$             \\ \hline
\multicolumn{3}{l|}{\multirow{2}{*}{B+HGFC+VAAB}} & 93.62             & 94.61             & 95.43             & 93.29             \\
\multicolumn{3}{l|}{}                                & $\textcolor{red}{\uparrow} \scriptsize{ \textcolor{red}{2.79}}$              & $\textcolor{red}{\uparrow} \scriptsize{ \textcolor{red}{1.76}}$              & $\textcolor{red}{\uparrow} \scriptsize{ \textcolor{red}{0.86}}$              & $\textcolor{red}{\uparrow} \scriptsize{ \textcolor{red}{2.42}}$              \\ \hline
\end{tabular}}
\end{table}

\begin{figure}[t]
\centering
	\begin{minipage}{0.48\linewidth}
		\centerline{\includegraphics[width=1.0\linewidth]{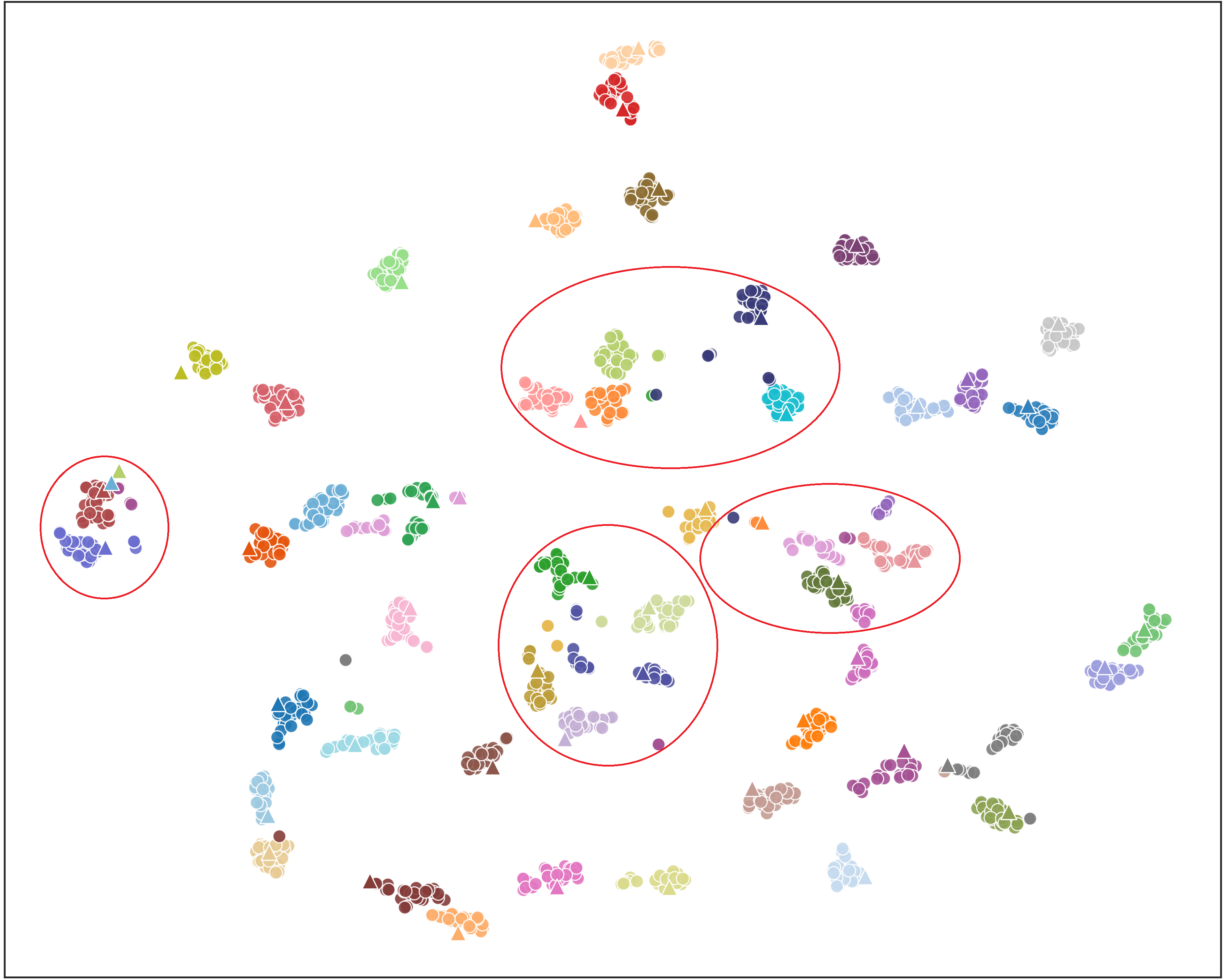}}
		\centerline{(a) Baseline}
	\end{minipage}
	\begin{minipage}{0.48\linewidth}
		\centerline{\includegraphics[width=1.0\linewidth]{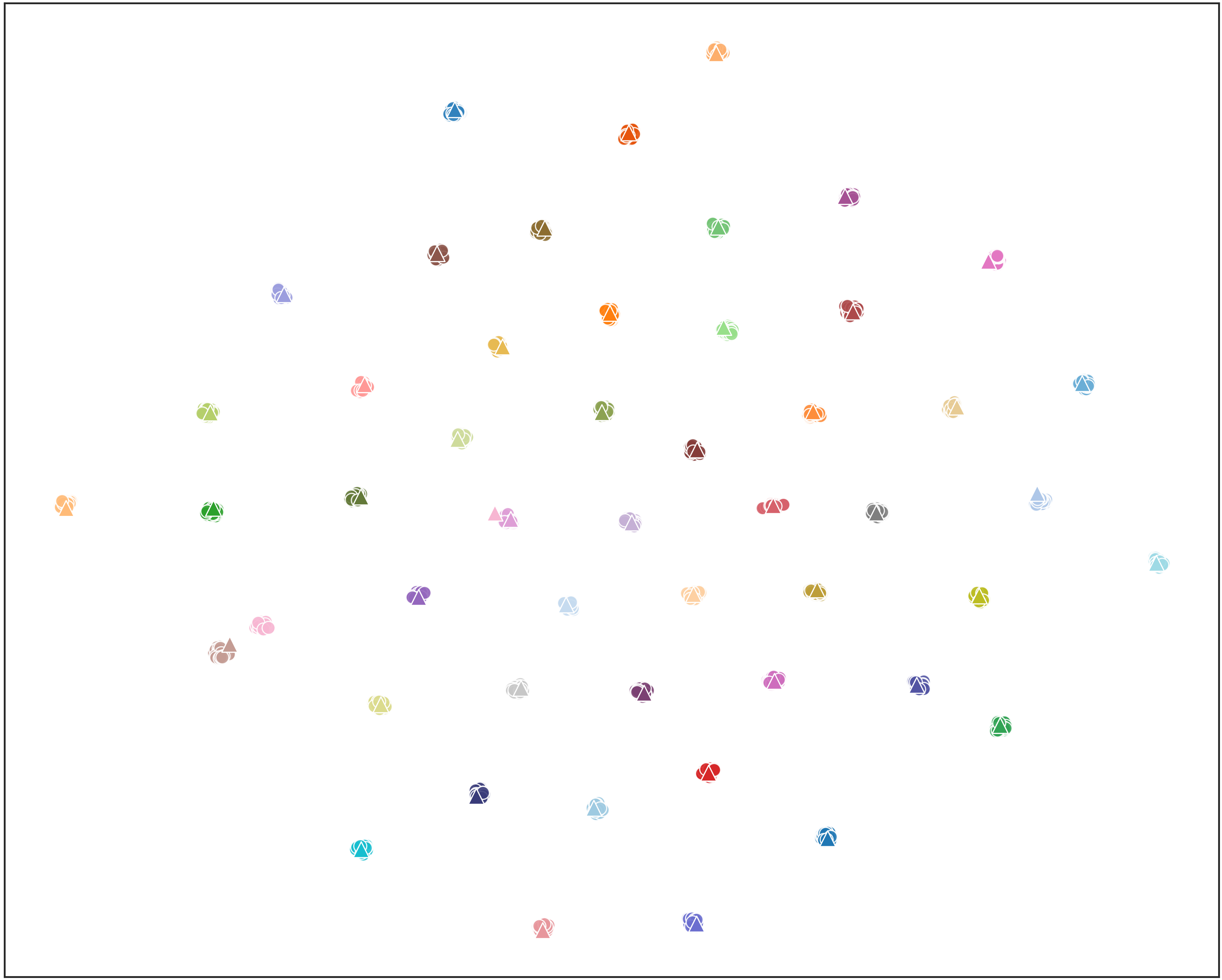}}
		\centerline{(b) Our}
	\end{minipage}
	\caption{The t-SNE visualization of the baseline and our proposed method. We randomly selected 50 location categories from the University-1652 dataset, with each color representing a location. ``\(\triangle\)'' represents satellite features, and ``\(\circ\)" represents UAV features. Red circles indicate ambiguous matches.}
\label{fig:vis_result}
\end{figure}

\begin{table*}[t]
 \renewcommand{\arraystretch}{1.25}
    \centering
    \caption{Performance (R@1 \% and AP \%) comparison with SOTA methods on the University-1652 dataset. Supervised (S); Unsupervised (U). Best results are marked in \textbf{bold}.}
    \label{tab:university_results}
     {\begin{tabular}{c|c|cc|cc}
\hline
\multirow{2}{*}{}  & \multirow{2}{*}{Method} & \multicolumn{2}{c|}{Drone → Satellite}               & \multicolumn{2}{c}{Satellite → Drone}                \\ \cline{3-6} 
                   &                         & \multicolumn{1}{c|}{R@1}            & AP             & \multicolumn{1}{c|}{R@1}            & AP             \\ \hline
\multirow{6}{*}{S} & Zhen et al. \cite{zheng2020university}            & \multicolumn{1}{c|}{59.69}          & 64.8           & \multicolumn{1}{c|}{73.18}          & 59.4           \\
                   & PCL \cite{tian2021uav}                    & \multicolumn{1}{c|}{79.47}          & 83.63          & \multicolumn{1}{c|}{87.69}          & 78.51          \\
                   & F3-net \cite{sun2023f3}                 & \multicolumn{1}{c|}{78.64}          & 81.60          & \multicolumn{1}{c|}{-}               & -               \\
                   & Sample4Geo \cite{deuser2023sample4geo}             & \multicolumn{1}{c|}{92.65}          & 93.81          & \multicolumn{1}{c|}{95.14}          & 91.39          \\
                   & DAC \cite{xia2024enhancing}                    & \multicolumn{1}{c|}{94.67} & 95.50 & \multicolumn{1}{c|}{96.43} & 93.79 \\ 
                   & QDFL \cite{hu2025query}             & \multicolumn{1}{c|}{\textbf{95.00}}          & \textbf{95.83}          & \multicolumn{1}{c|}{\textbf{97.15}}          & \textbf{94.57}          \\ \hline
\multirow{3}{*}{U} & Li et al. \cite{li2024learning}              & \multicolumn{1}{c|}{70.29}          & 74.93          & \multicolumn{1}{c|}{79.03}          & 61.03          \\
                   & Wang et al. \cite{wang2025coarse}            & \multicolumn{1}{c|}{85.95}          & 90.33          & \multicolumn{1}{c|}{94.01}          & 82.66          \\ 
                   & UniABG                  & \multicolumn{1}{c|}{\textbf{93.62}} & \textbf{94.61} & \multicolumn{1}{c|}{\textbf{95.43}} & \textbf{93.29} \\ \hline
\end{tabular}}
\end{table*}

\begin{table*}[!ht]
 \renewcommand{\arraystretch}{1.25}
    \centering
    \caption{Performance (R@1 \% and AP \%) comparison with SOTA methods on the SUES-200 dataset. Supervised (S); Unsupervised (U). Best results are marked in \textbf{bold}.}
    \label{tab:sues_results}
     {
     \begin{tabular}{cccccccccc}
\hline
\multicolumn{10}{c}{Drone → Satellite}                                                                                                                                                                                                                                                                                                                                             \\ \hline
\multicolumn{1}{c|}{\multirow{2}{*}{}}  & \multicolumn{1}{c|}{\multirow{2}{*}{Method}} & \multicolumn{2}{c|}{150m}                                                 & \multicolumn{2}{c|}{200m}                                                 & \multicolumn{2}{c|}{250m}                                                  & \multicolumn{2}{c}{300m}                             \\ \cline{3-10} 
\multicolumn{1}{c|}{}                   & \multicolumn{1}{c|}{}                        & \multicolumn{1}{c|}{R@1}            & \multicolumn{1}{c|}{AP}             & \multicolumn{1}{c|}{R@1}            & \multicolumn{1}{c|}{AP}             & \multicolumn{1}{c|}{R@1}             & \multicolumn{1}{c|}{AP}             & \multicolumn{1}{c|}{R@1}            & AP             \\ \hline
\multicolumn{1}{c|}{\multirow{5}{*}{S}} & \multicolumn{1}{c|}{SUES-200 \cite{zhu2023sues}}                & \multicolumn{1}{c|}{55.65}          & \multicolumn{1}{c|}{61.92}          & \multicolumn{1}{c|}{66.78}          & \multicolumn{1}{c|}{71.55}          & \multicolumn{1}{c|}{72.00}           & \multicolumn{1}{c|}{76.43}          & \multicolumn{1}{c|}{74.05}          & 78.26          \\
\multicolumn{1}{c|}{}                   & \multicolumn{1}{c|}{FSRA \cite{dai2021transformer}}                    & \multicolumn{1}{c|}{68.25}          & \multicolumn{1}{c|}{73.45}          & \multicolumn{1}{c|}{83.00}          & \multicolumn{1}{c|}{85.99}          & \multicolumn{1}{c|}{90.68}           & \multicolumn{1}{c|}{92.27}          & \multicolumn{1}{c|}{91.95}          & 93.46          \\
\multicolumn{1}{c|}{}                   & \multicolumn{1}{c|}{MCCG \cite{shen2023mccg}}                    & \multicolumn{1}{c|}{82.22}          & \multicolumn{1}{c|}{85.47}          & \multicolumn{1}{c|}{89.38}          & \multicolumn{1}{c|}{91.41}          & \multicolumn{1}{c|}{93.82}           & \multicolumn{1}{c|}{95.04}          & \multicolumn{1}{c|}{95.07}          & 96.20          \\
\multicolumn{1}{c|}{}                   & \multicolumn{1}{c|}{DAC \cite{xia2024enhancing}}                     & \multicolumn{1}{c|}{\textbf{96.80}} & \multicolumn{1}{c|}{\textbf{97.54}} & \multicolumn{1}{c|}{97.48} & \multicolumn{1}{c|}{97.97} & \multicolumn{1}{c|}{98.20}  & \multicolumn{1}{c|}{98.62} & \multicolumn{1}{c|}{97.58} & 98.14 \\
\multicolumn{1}{c|}{}                   & \multicolumn{1}{c|}{QDFL \cite{hu2025query}}                     & \multicolumn{1}{c|}{93.97} & \multicolumn{1}{c|}{95.42} & \multicolumn{1}{c|}{\textbf{98.25}} & \multicolumn{1}{c|}{\textbf{98.67}} & \multicolumn{1}{c|}{\textbf{99.30}}  & \multicolumn{1}{c|}{\textbf{99.48}} & \multicolumn{1}{c|}{\textbf{99.31}} & \textbf{99.48} \\ \hline
\multicolumn{1}{c|}{\multirow{2}{*}{U}} & \multicolumn{1}{c|}{Wang et al. \cite{wang2025coarse}}             & \multicolumn{1}{c|}{76.90}          & \multicolumn{1}{c|}{84.95}          & \multicolumn{1}{c|}{87.88}          & \multicolumn{1}{c|}{92.60}          & \multicolumn{1}{c|}{92.98}           & \multicolumn{1}{c|}{95.66}          & \multicolumn{1}{c|}{95.10}          & 96.92          \\
\multicolumn{1}{c|}{}                   & \multicolumn{1}{c|}{UniABG}                  & \multicolumn{1}{c|}{\textbf{92.40}} & \multicolumn{1}{c|}{\textbf{93.95}} & \multicolumn{1}{c|}{\textbf{97.32}} & \multicolumn{1}{c|}{\textbf{97.92}} & \multicolumn{1}{c|}{\textbf{98.07}}  & \multicolumn{1}{c|}{\textbf{98.55}} & \multicolumn{1}{c|}{\textbf{98.67}} & \textbf{98.98} \\ \hline
\multicolumn{10}{c}{Satellite → Drone}                                                                                                                                                                                                                                                                                                                                             \\ \hline
\multicolumn{1}{c|}{\multirow{2}{*}{}}  & \multicolumn{1}{c|}{\multirow{2}{*}{Method}} & \multicolumn{2}{c|}{150m}                                                 & \multicolumn{2}{c|}{200m}                                                 & \multicolumn{2}{c|}{250m}                                                  & \multicolumn{2}{c}{300m}                             \\ \cline{3-10} 
\multicolumn{1}{c|}{}                   & \multicolumn{1}{c|}{}                        & \multicolumn{1}{c|}{R@1}            & \multicolumn{1}{c|}{AP}             & \multicolumn{1}{c|}{R@1}            & \multicolumn{1}{c|}{AP}             & \multicolumn{1}{c|}{R@1}             & \multicolumn{1}{c|}{AP}             & \multicolumn{1}{c|}{R@1}            & AP             \\ \hline
\multicolumn{1}{c|}{\multirow{5}{*}{S}} & \multicolumn{1}{c|}{SUES-200 \cite{zhu2023sues}}                & \multicolumn{1}{c|}{75.00}          & \multicolumn{1}{c|}{55.46}          & \multicolumn{1}{c|}{85.00}          & \multicolumn{1}{c|}{66.05}          & \multicolumn{1}{c|}{86.25}           & \multicolumn{1}{c|}{69.94}          & \multicolumn{1}{c|}{88.75}          & 74.46          \\
\multicolumn{1}{c|}{}                   & \multicolumn{1}{c|}{FSRA \cite{dai2021transformer}}                    & \multicolumn{1}{c|}{83.75}          & \multicolumn{1}{c|}{76.67}          & \multicolumn{1}{c|}{90.00}          & \multicolumn{1}{c|}{85.34}          & \multicolumn{1}{c|}{93.75}           & \multicolumn{1}{c|}{90.17}          & \multicolumn{1}{c|}{95.00}          & 92.03          \\
\multicolumn{1}{c|}{}                   & \multicolumn{1}{c|}{MCCG \cite{shen2023mccg}}                    & \multicolumn{1}{c|}{97.50}          & \multicolumn{1}{c|}{93.63}          & \multicolumn{1}{c|}{98.75}          & \multicolumn{1}{c|}{96.70}          & \multicolumn{1}{c|}{98.75}           & \multicolumn{1}{c|}{98.28}          & \multicolumn{1}{c|}{98.75}          & 98.05          \\
\multicolumn{1}{c|}{}                   & \multicolumn{1}{c|}{DAC \cite{xia2024enhancing}}                     & \multicolumn{1}{c|}{97.50} & \multicolumn{1}{c|}{94.06} & \multicolumn{1}{c|}{\textbf{98.75}} & \multicolumn{1}{c|}{96.66} & \multicolumn{1}{c|}{98.75}  & \multicolumn{1}{c|}{98.09} & \multicolumn{1}{c|}{98.75} & 97.87 \\ 
\multicolumn{1}{c|}{}                   & \multicolumn{1}{c|}{QDFL \cite{hu2025query}}                     & \multicolumn{1}{c|}{\textbf{98.75}} & \multicolumn{1}{c|}{\textbf{95.10}} & \multicolumn{1}{c|}{\textbf{98.75}} & \multicolumn{1}{c|}{\textbf{97.92}} & \multicolumn{1}{c|}{\textbf{100.00}}  & \multicolumn{1}{c|}{\textbf{99.07}} & \multicolumn{1}{c|}{\textbf{100.00}} & \textbf{99.07} \\ \hline
\multicolumn{1}{c|}{\multirow{2}{*}{U}} & \multicolumn{1}{c|}{Wang et al. \cite{wang2025coarse}}             & \multicolumn{1}{c|}{87.50}          & \multicolumn{1}{c|}{74.81}          & \multicolumn{1}{c|}{92.50}          & \multicolumn{1}{c|}{87.15}          & \multicolumn{1}{c|}{96.25}           & \multicolumn{1}{c|}{91.20}          & \multicolumn{1}{c|}{\textbf{98.75}}          & 94.52          \\
\multicolumn{1}{c|}{}                   & \multicolumn{1}{c|}{UniABG}                  & \multicolumn{1}{c|}{\textbf{98.75}} & \multicolumn{1}{c|}{\textbf{91.54}} & \multicolumn{1}{c|}{\textbf{98.75}} & \multicolumn{1}{c|}{\textbf{97.06}} & \multicolumn{1}{c|}{\textbf{100.00}} & \multicolumn{1}{c|}{\textbf{98.32}} & \multicolumn{1}{c|}{\textbf{98.75}} & \textbf{97.58} \\ \hline
\end{tabular}}
\end{table*}

\subsection{Ablation Study} 
We ablate two core components: View-Aware Adversarial Bridging (VAAB) and Heterogeneous Graph Filtering Calibration (HGFC). These progressively mitigate cross-view variations while enhancing association quality. Table \ref{tab:ablation_results} quantifies contributions under identical settings:
1) \textit{Effect of the HGFC:} HGFC enforces structural consistency for robust associations. For Drone $\rightarrow$ Satellite, it boosts R@1 by +54.89\% and AP by +51.21\%. For Satellite $\rightarrow$ Drone, gains reach +29.1\% R@1 and +55.26\% AP. This verifies that structural filtering is essential to overcome ambiguous appearance-based matches.
2) \textit{Effect of the VAAB:} VAAB reduces domain gaps through adversarial pseudo-view (APV) generation. It yields further improvements: +2.79\% R@1 / +1.76\% AP for Drone $\rightarrow$ Satellite and +0.86\% R@1 / +2.42\% AP for Satellite $\rightarrow$ Drone. This confirms that explicit view-invariant learning via APV is critical for feature robustness. 3) \textit{Visual Analysis:} 
Fig. \ref{fig:vis_result} compares t-SNE visualizations of baseline and our UniABG. Baseline features show poor clustering with incorrect matches. Our model tightly clusters cross-view features of the same class while separating different classes with distinct margins.

\subsection{Comparison with the State-of-the-art Methods} 
We evaluate our method on University-1652 and SUES-200 benchmarks. On University-1652 (Table \ref{tab:university_results}), our approach advances the state-of-the-art by significant margins: +7.67\% Drone $\rightarrow$ Satellite R@1 and +10.63\% Satellite $\rightarrow$ Drone AP. 

SUES-200 results (Table \ref{tab:sues_results}) further demonstrate superiority, with particularly notable gains at the challenging 150m altitude. For Drone $\rightarrow$ Satellite retrieval, we achieve +15.5\% R@1 and +9.0\% AP. In the Satellite $\rightarrow$ Drone direction, improvements reach +11.25\% R@1 and +16.73\% AP.

\section{Conclusion} 
We propose UniABG, a dual-stage unsupervised framework addressing viewpoint discrepancy and association noise in cross-view geo-localization. Specifically, our approach integrates adversarial learning to align drone-satellite feature distributions, while simultaneously employing heterogeneous graph filtering to resolve ambiguous matches and suppress error propagation during association construction. Experimental results demonstrate state-of-the-art unsupervised performance on University-1652 and SUES-200 benchmarks. Crucially, UniABG establishes that combining structural graph filtering with adversarial learning generates robust feature representations for label-free localization. Collectively, this work provides both an effective solution for cross-view geo-localization and a promising paradigm for future cross-modal matching research.

\section*{Acknowledgments}
This work is partially supported by National Natural Science Foundation of China under Grants (62306215, 62501428),  Postdoctoral Fellowship Program of China Postdoctoral Science Foundation (GZC20241268, 2024M762479),  Hubei Postdoctoral Talent Introduction Program (2024HBBHJD070) and Hubei Provincial Natural Science Foundation of China (2025AFB219).The numerical calculations in this paper had been supported by the super-computing system in the Supercomputing Center of Wuhan University.


\section{Appendices}
To further enhance the reproducibility and provide deeper insights into our design choices, we analyze two key hyperparameters in our framework: (1) the number of nearest neighbors $k$ used in heterogeneous graph filtering, and (2) the adversarial weight $\lambda$ in Stage~1. These parameters play critical roles in refining cross-view associations and balancing view-invariant alignment with intra-view discrimination. The following subsections present detailed discussions and ablation results for $k$ and $\lambda$.

\begin{table}[h]
    \centering
    \caption{Impact of the number of neighbors $k$ on performance.}
    \label{tab:k_neighbors}
    {
    \begin{tabular}{l|llll}
\hline
\multirow{2}{*}{k} & \multicolumn{2}{l}{Drone→Satellite} & \multicolumn{2}{l}{Satellite→Drone} \\
                   & R@1              & AP               & R@1              & AP               \\ \hline
1                  & 93.75            & 94.68            & 96.29            & 93.57            \\
2                  & 93.62            & 94.61            & 95.43            & 93.29            \\
3                  & 93.07            & 94.08            & 95.86            & 93.10            \\
4                  & 93.00            & 94.07            & 96.14            & 93.25            \\ \hline
\end{tabular}}
\end{table}

\subsection{Sensitivity Analysis: Impact of the Number of Neighbors $k$}
To further validate the robustness of our proposed Heterogeneous Graph Filtering Calibration (HGFC) module, we conduct a sensitivity analysis on the number of nearest neighbors $k$ used for cross-view graph construction. Specifically, we vary $k$ from 1 to 4 and report Recall@1 (R@1) and Average Precision (AP) for both Drone$\rightarrow$Satellite and Satellite$\rightarrow$Drone retrieval tasks, as shown in Table~\ref{tab:k_neighbors}.

The results indicate that the performance of our method remains highly stable across different values of $k$, with variations in R@1 and AP within $\pm$0.75\%. This demonstrates that our approach is insensitive to the choice of $k$, exhibiting strong robustness. Such stability can be attributed to the mutual $k$-reciprocal filtering mechanism in HGFC, which effectively preserves structurally consistent matches while suppressing noisy neighbors.

Notably, the best results are achieved when $k=1$ or $k=2$, but even with larger $k$, the performance degradation is negligible. This confirms that our heterogeneous graph filtering strategy reliably constructs high-quality cross-view associations under different neighborhood sizes, significantly mitigating the risk of hyperparameter sensitivity.

In summary, this experiment validates that \textbf{UniABG} maintains stable performance across varying $k$, highlighting its robustness and practicality for unsupervised cross-view geo-localization without requiring extensive hyperparameter tuning.


\begin{table}[h]
    \centering
    \caption{Effect of the hyperparameter $\lambda$ on University-1652.}
    \label{tab:lambda}
    {
   \begin{tabular}{c|cccc}
\hline
\multirow{2}{*}{$\lambda$} & \multicolumn{2}{c}{Drone → Satellite} & \multicolumn{2}{c}{Satellite → Drone} \\
                   & R@1               & AP                & R@1               & AP                \\ \hline
0.1                & 93.62             & 94.61             & 95.43             & 93.29             \\
0.2                & 93.60             & 94.57             & 96.29             & 93.62             \\
0.3                & 93.22             & 94.24             & 95.57             & 92.84             \\
0.4                & 92.78             & 93.88             & 95.14             & 92.81             \\
0.5                & 93.46             & 94.44             & 96.29             & 93.52             \\
0.6                & 93.18             & 94.22             & 95.14             & 92.92             \\
0.7                & 92.60             & 93.68             & 95.43             & 92.77             \\
0.8                & 92.27             & 93.45             & 95.00             & 92.00             \\
0.9                & 92.15             & 93.30             & 95.86             & 92.33             \\
1.0                & 91.90             & 93.17             & 95.72             & 92.22             \\ \hline
\end{tabular}}
\end{table}

\subsection{Discussion on Hyperparameter $\lambda$}

In our framework, the hyperparameter $\lambda$ balances the intra-view contrastive loss $L_{iv}$ and the view-aware adversarial loss $L_{\text{VAAB}}$ in Stage~1, as defined in Eq.~(24). To investigate its effect, we vary $\lambda$ from 0.1 to 1.0 and report the performance on the University-1652 benchmark, as shown in Table~\ref{tab:lambda}.

As shown in Table~\ref{tab:lambda}, $\lambda = 0.1$ yields the best overall performance for both \emph{Drone $\rightarrow$ Satellite} and \emph{Satellite $\rightarrow$ Drone}, achieving 93.62\% / 95.43\% in R@1 and 94.61\% / 93.29\% in AP, respectively. Increasing $\lambda$ beyond this point consistently leads to performance degradation. This indicates that over-emphasizing the adversarial loss disrupts the discriminability of intra-view features learned by $L_{iv}$, resulting in unstable clustering and noisier cross-view associations.

On the other hand, too small a value weakens the adversarial bridging effect, leaving the domain gap insufficiently mitigated and causing performance drops. Therefore, $\lambda = 0.1$ strikes a desirable balance between enforcing view-invariant alignment and maintaining intra-view structural discrimination. This balanced setting ensures that Stage~1 produces robust embeddings, which in turn facilitates high-quality graph-based correspondence refinement in Stage~2. These findings are consistent with prior adversarial alignment studies, confirming that moderate adversarial regularization is crucial for stabilizing unsupervised cross-view geo-localization.


\subsection{Elaboration on the Auxiliary Pseudo View (APV)}

 This section provides a comprehensive elaboration on the Auxiliary Pseudo View (APV), clarifying its motivation, generation process, and role within our framework. A primary challenge in unsupervised cross-view geo-localization is the significant domain gap between drone-view and satellite-view imagery, which differ not only in perspective but also in global illumination, color palettes, and spectral properties. The APV is designed explicitly to address this challenge by serving as a "view bridge." The core idea is to create a synthetic, intermediate domain that semantically lies between the drone and satellite views. By compelling our feature extractor to learn representations that are invariant across all three domains—the original drone view, the generated APV, and the satellite view—the model is guided to disentangle domain-specific variations from essential, location-discriminative features. This process effectively reduces the domain discrepancy, leading to more reliable pseudo-label generation and ultimately enhancing matching accuracy.

 To generate the APV, we employ the classic yet computationally efficient color transfer algorithm proposed by Reinhard et al. (2001). This method was chosen for its effectiveness in matching global color statistics without requiring complex network training, making it a lightweight and stable component. The process involves converting the source drone image and target satellite images to the decorrelated $l\alpha\beta$ color space to separate intensity from color information. We then pre-calculate the global mean and standard deviation for each channel across the entire satellite dataset to define the target style. For each drone image, its color statistics are normalized and scaled to match these global satellite statistics. The resulting image, after being converted back to the BGR color space, retains the geometric structure of the original drone view while adopting the global color and illumination characteristics of the satellite view. To visually demonstrate this effect, Figure S1 provides a side-by-side comparison of original drone views, their corresponding generated APVs, and the ground-truth satellite views, clearly showing that the APV successfully bridges the visual gap.

\begin{figure}[H] 
    \centering 
    \includegraphics[width=0.45\textwidth]{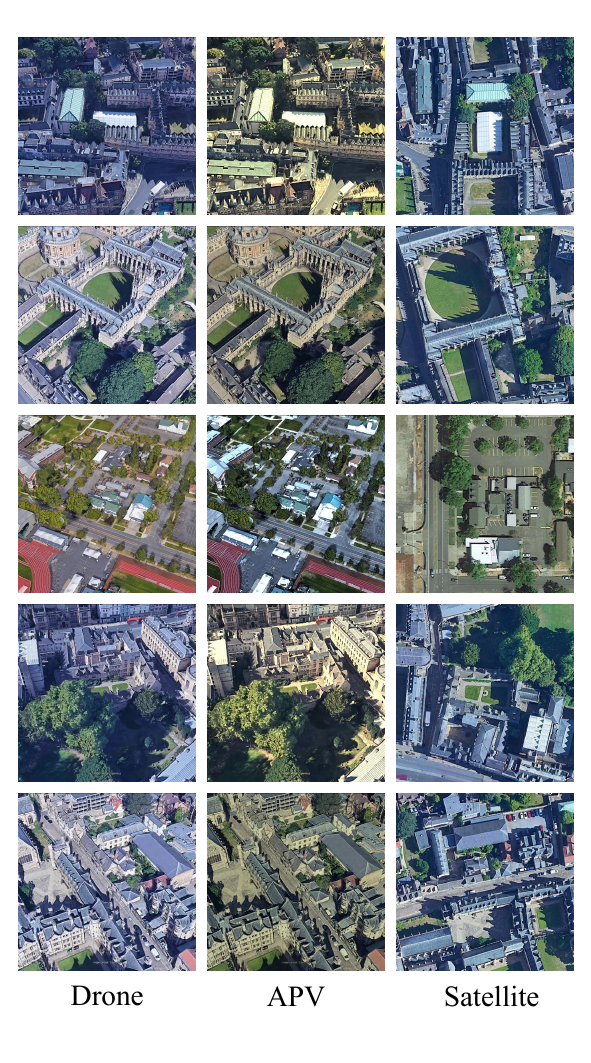} 
    \caption{Visual examples of the generated Auxiliary Pseudo View (APV). For each sample, the left column shows the original drone-view image. The middle column displays the generated APV after applying color transfer. The right column presents the corresponding ground-truth satellite-view image.} 
    \label{fig:APV_Vis} 
\end{figure}

 Figure \ref{fig:APV_Vis}: Visual examples of the generated Auxiliary Pseudo View (APV). For each sample, the left column shows the original drone-view image. The middle column displays the generated APV after applying color transfer. The right column presents the corresponding ground-truth satellite-view image. It is evident that the APV successfully preserves the structural and geometric layout of the drone view while emulating the global color characteristics of the satellite imagery. This transformation effectively reduces the visual domain gap, thereby acting as a "view bridge" that facilitates more robust feature learning.

 Our choice of this method provides a robust and efficient mechanism for creating the APV, proving the effectiveness of our view-bridging concept. As noted in our response to the reviewers, exploring more advanced style transfer techniques is a promising direction for future work. Methods based on Generative Adversarial Networks (GANs), such as CycleGAN, could potentially generate even more realistic APVs by capturing more complex, non-linear texture and style variations. Such advancements might further close the domain gap and yield additional performance improvements. However, our current approach validates the core contribution of using an APV as a foundational strategy and provides a strong balance between performance and computational efficiency.

\section{Dual-Stage Training Framework}
We summarize our full training pipeline combining Stage~1 (View-Aware Adversarial Bridging) and Stage~2 (Cross-View Association Learning) as follows.

\begin{algorithm}[h]
\caption{Dual-Stage Training Framework}
\label{alg:dual_stage}
\begin{algorithmic}[1]
\REQUIRE Satellite loader $\mathcal{D}_s$, Drone loader $\mathcal{D}_d$, optimizer $\theta$, view classifier optimizer $\theta_v$
\ENSURE Updated encoder $\mathcal{F}_{\text{backbone}}$ and view classifier $\mathcal{D}_v$

\STATE \textbf{Stage 1: Intra-view Contrastive Learning + View-Aware Adversarial Bridging}
\FOR{$epoch = 1$ to $E_1$}
    \FOR{$iter = 1$ to $T_1$}
        \STATE \textit{(1) Load satellite and drone batches}
        \STATE $\mathbf{x}_s \leftarrow \mathcal{D}_s.\text{next()}$, $\mathbf{x}_d \leftarrow \mathcal{D}_d.\text{next()}$
        \STATE \textit{(2) Feature extraction}
        \STATE $\mathbf{f}^s, \mathbf{f}^d \leftarrow \mathcal{F}(\mathbf{x}_s), \mathcal{F}(\mathbf{x}_d)$
        \STATE \textit{(3) Memory-based intra-view contrastive loss (Eq.~8--9)}
        \STATE $L_{sat} \leftarrow \mathcal{M}_s(\mathbf{f}_s)$,\quad $L_{drone} \leftarrow \mathcal{M}_d(\mathbf{f}_d)$
        \IF{$epoch \leq 1$}
            \STATE $L \leftarrow L_{sat} + L_{drone}$
        \ELSE
            \STATE \textit{(4) Update view classifier (Eq.~14 first term)}
            \STATE $L_{\text{view}} \leftarrow \text{CE}(\mathcal{D}_v(\mathbf{f}^v), t^v)$
            \STATE $\theta_v.\text{step}(L_{\text{view}})$
            \STATE \textit{(5) Adversarial bridging loss (Eq.~14 second term)}
            \STATE $L_{\text{adv}} \leftarrow \text{CE}(\mathcal{D}_v(\mathbf{f}^v), t^p)$
            \STATE \textit{(6) Joint optimization}
            \STATE $L \leftarrow L_{sat} + L_{drone} + \lambda L_{\text{adv}}$
        \ENDIF
        \STATE $\theta.\text{step}(L)$
    \ENDFOR
\ENDFOR

\STATE \textbf{Stage 2: Cross-View Association Learning}
\FOR{$epoch = 1$ to $E_2$}
    \FOR{each batch $(q, r, y)$}
        \STATE \textit{(1) Forward pass with mixed precision}
        \STATE Move $q, r, y$ to device
        \STATE Extract features: $\{f_{ctr}, f_{cls}, f_{dsa}\} \leftarrow \mathcal{M}(q, r)$
        \STATE \textit{(2) InfoNCE contrastive loss (Eq.~11)}
        \STATE $L_{\text{InfoNCE}} \leftarrow \mathcal{L}_{\text{InfoNCE}}(f_{ctr}^q, f_{ctr}^r)$
        \STATE \textit{(3) Classification loss (Eq.~11)}
        \STATE $L_{\text{CE}} \leftarrow \text{CE}(f_{cls}^q, y) + \text{CE}(f_{cls}^r, y)$
        \STATE \textit{(4) Domain Space Alignment loss (Eq.~11)}
        \STATE $L_{\text{MSE}} \leftarrow \frac{1}{N} \sum_{i=1}^N \| f_{dsa,i}^q - f_{dsa,i}^r \|_2^2$
        \STATE \textit{(5) Joint optimization}
        \STATE $L_{\text{Stage2}} \leftarrow L_{\text{InfoNCE}} + L_{\text{MSE}} + L_{\text{CE}}$
        \STATE $\theta.\text{step}(L_{\text{Stage2}})$
    \ENDFOR
\ENDFOR
\end{algorithmic}
\end{algorithm}

The overall objective is defined as:
\begin{equation}
L = 
\begin{cases}
L_{iv} + \lambda L_{\text{VAAB}}, & \text{Stage 1 (Eq.~24)} \\
L_{\text{InfoNCE}} + L_{\text{MSE}} + L_{\text{CE}}, & \text{Stage 2 (Eq.~11)}
\end{cases}
\end{equation}

\end{document}